\documentclass[runningheads]{llncs} 
\usepackage[utf8]{inputenc}
\usepackage{amsmath,amssymb,amsfonts}
\usepackage{graphicx}
\usepackage{booktabs}
\usepackage{multirow}
\usepackage{xcolor}
\usepackage{hyperref}
\usepackage{algorithm}
\usepackage{algorithmic}
\usepackage{cite}
\usepackage{microtype}
\usepackage{caption}
\usepackage{subcaption}
\usepackage{float}

\hypersetup{
    colorlinks=true,
    linkcolor=blue,
    citecolor=blue,
    urlcolor=blue,
}

\title{\textbf{CMDiff: Left Atrial Scar Segmentation from 3D LGE MRI Using Conditional Mask Diffusion Model: A Feasibility Study}}
\titlerunning{LA Scar Segmentation Using Conditional Mask Diffusion Model}
\author{
    Jing Zhang\inst{1}\quad\quad
    Bastien Bergere\inst{1}\quad\quad
    Jonas Leite\inst{1}\quad\quad
    Emilie Bollache\inst{1}\quad\quad
    Mikaël Laredo\inst{1,2}\quad\quad
    Alban Redheuil\inst{1,3}\quad\quad
    Nadjia Kachenoura\inst{1}
}
\authorrunning{J. Zhang et al.}

\institute{
    Laboratoire d'Imagerie Biomédicale (LIB), Sorbonne Université, INSERM, CNRS, Paris, France
    \and
    APHP, Institut de Cardiologie, Hôpital Pitié-Salpêtrière, Paris, France
    \and 
    APHP, Imagerie Cardiovasculaire et Thoracique, Institut de Cardiologie, Hôpital Pitié-Salpêtrière, Paris, France
}

\begin{document}
\maketitle

\begin{abstract}
Late gadolinium enhancement (LGE) cardiac MRI enables non-invasive assessment of left atrial (LA) scar, an imaging marker associated with atrial fibrillation ablation outcome. However, automatic scar segmentation is hindered by low contrast, the thin atrial wall, scarce labels, and extreme class imbalance.
Conditional diffusion offers iterative mask generation, but its suitability for sparse 3D LA scar segmentation remains unclear.
To address this gap, we propose Conditional Mask Diffusion (CMDiff), which predicts clean masks using a dynamic-encoder-conditioned 3D sinusoidal UNet. We compared CMDiff with a deterministic baseline on identical cases in five-fold cross-validation of 60 scar-labelled public LASCARQS subjects, with 48 training and 12 held-out cases per fold.
CMDiff achieved a scar DSC of 0.47 $\pm$ 0.08, HD95 of 12.89 $\pm$ 6.39 mm, and ASD of 1.91 $\pm$ 0.96 mm, compared with deterministic baseline values of 0.51 $\pm$ 0.08, 10.86 $\pm$ 5.39 mm, and 1.58 $\pm$ 1.03 mm, respectively. Under the same implementation, clean-mask prediction performed substantially better than the noise-prediction configuration evaluated in this study.
Thus, CMDiff was technically feasible but did not improve segmentation performance: DSC was lower, and both HD95 and ASD were higher. Direct mask diffusion therefore provides an analytical alternative, but not a competitive performance advance for this task. The source code is publicly available at \url{https://github.com/jizhang02/CMDiff}.
\end{abstract}


\newpage
\section{Introduction}
\label{sec:intro}

Atrial fibrillation (AF) is the most prevalent sustained cardiac arrhythmia, affecting an estimated 37.6 million people worldwide and increasing the risks of stroke, heart failure, and mortality~\cite{dai2021global}. Catheter ablation is an established rhythm-control therapy for symptomatic AF, but recurrence remains common. Because the burden and spatial distribution of left atrial (LA) fibrosis have been associated with ablation outcome~\cite{marrouche2014association}, late gadolinium enhancement (LGE) cardiac MRI may help characterise the atrial substrate before treatment and scar after treatment~\cite{peters2009recurrence}.

Converting LGE-MRI into reliable scar maps is nevertheless difficult. Scar-related signal is subtle and confined to the thin LA wall ($\sim$2--4\,mm), where it can resemble enhancement from the blood pool or adjacent tissue. Scar voxels are also sparse, manual boundaries vary between observers, and image appearance differs across patients, scanners, and acquisition protocols~\cite{li2022medical}. These constraints motivate segmentation methods that combine local LGE appearance with explicit anatomical context.

Early approaches treated scar mainly as an intensity outlier, using thresholding or standardised intensity rules~\cite{oakes2009detection,karim2014}. Convolutional encoder--decoder models then learned appearance and volumetric context directly from images~\cite{ronneberger2015unet,cciccek20163d,he2016deep}, while nnU-Net provided a strong self-configuring baseline~\cite{isensee2021nnu}. Cardiac-specific methods further combined learned scar likelihoods with graph-cut regularisation, multi-view information, attention, spatial encoding, or joint LA--scar objectives~\cite{li2020b,li2020atrialnet,chen2018b,yang2020,li2020a,li2022atrialjsqnet,wu2022multidepth}.

More recent methods increasingly constrain scar to the LA wall. A boundary-focused nnU-Net used the predicted LA distance map and reported a scar DSC of 64.08\% on public LASCARQS dataset~\cite{zhang2023boundary}; an anatomy-conditioned two-stage pipeline combined cavity and wall signed-distance maps with wall-restricted, boundary-aware supervision and reported 61.1\%~\cite{kundu2026twostage}. Our prior work similarly used a coarse LA prediction to constrain SwinUNETR-based scar delineation~\cite{zhang2026progressive,liu2021swin,hatamizadeh2021swin}. These values are contextual rather than directly comparable because data partitions, model selection, and evaluation implementations may differ. Collectively, however, they show that deterministic performance depends strongly on how anatomy is incorporated.

These anatomy-aware models remain deterministic: each image produces one mask. Diffusion models offer a different formulation by reversing a gradual noising process from different noise initialisations~\cite{ho2020ddpm}. Applied to segmentation, the mask is denoised while the medical image supplies the condition. This principle has supported implicit segmentation ensembles~\cite{wolleb2022diffusion}, noise-prediction and dynamic-conditioning models~\cite{wu2024medsegdiff,wu2024medsegdiffv2}, and distributions of plausible masks learned from multiple expert annotations~\cite{rahman2023ambiguous}. 

Within cardiac imaging, diffusion has been used to enhance left-ventricular scar contrast~\cite{hasny2024localized} and generate synthetic atrial fibrosis distributions~\cite{zolotarev2025synthetic}, rather than segment patient-specific LA scar masks. Whether a 3D mask-denoising process can recover LA scar directly from LGE-MRI therefore remains unresolved.

To address this gap, we study diffusion as an alternative rather than a presumed replacement for deterministic segmentation. We test whether 3D Conditional Mask Diffusion (CMDiff) can approach a matched deterministic baseline, identify effective and ineffective design choices under a common implementation, and describe repeated-sample variability without treating it as calibrated uncertainty. CMDiff predicts the clean mask at each denoising step to align the diffusion output with the sparse binary target.

This paper makes the following contributions:
\vspace{-0.1cm}
\begin{enumerate}
    \item We present CMDiff, to our knowledge the first task-specific feasibility and failure-mode study of 3D LA scar segmentation formulated as mask diffusion conditioned on the LGE image.
    \item We compare clean-mask and noise-prediction parameterisations under the same implementation and show that only clean-mask prediction produces useful scar masks in this experimental configuration.
    \item We use a lightweight Dynamic Conditional Encoder (DCE) to inject multi-scale LGE features into the denoising process.
    \item We conduct target, conditioning, and sampler ablations; compare CMDiff with a specialised deterministic LA scar segmentation model on identical held-out cases; and describe stochastic sample disagreement through entropy and variance maps.
\end{enumerate}

\section{Methods}
\label{sec:methods}

\begin{figure}[t]
    \centering
    \includegraphics[width=0.95\linewidth]{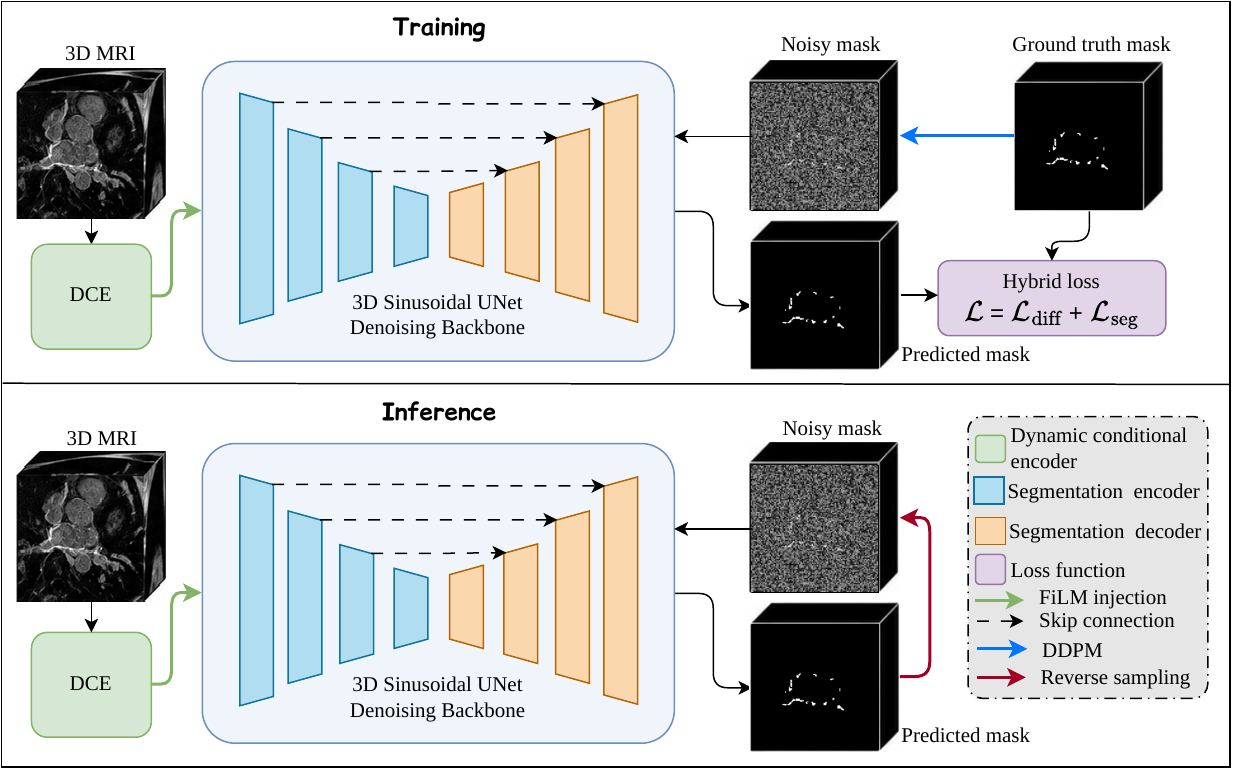}
    \caption{Overview of the proposed CMDiff framework for 3D LGE-MRI scar segmentation. During training, noise is added to the ground-truth mask and the denoising backbone learns to recover the clean scar mask conditioned on the LGE image. During inference, a noisy mask is iteratively denoised through the sampling scheduler, while the DCE provides multi-scale image conditioning to the 3D sinusoidal UNet.}
    \label{fig:diffseg_model}
\end{figure}

CMDiff treats the scar mask, rather than the LGE image, as the variable in a conditional diffusion process. During training, Gaussian noise is added to the binary mask and the network learns to recover its clean form from the noisy mask, timestep, and LGE image. During inference, the learned reverse process transforms a random mask into a segmentation. Figure~\ref{fig:diffseg_model} summarises this workflow; the following subsections define the diffusion target, image conditioning, and optimisation objective.

\subsection{Conditional mask diffusion model (CMDiff)}
Let $\boldsymbol{x} \in \mathbb{R}^{D \times H \times W}$ denote a 3D LGE-MRI patch and $\boldsymbol{y} \in \{0,1\}^{D \times H \times W}$ the corresponding binary scar mask.
Following DDPMs~\cite{ho2020ddpm}, we learn a denoising model $g_\theta$ that reverses a gradual Gaussian noising process.
To make the binary mask compatible with zero-centred Gaussian diffusion, we map $\boldsymbol{y}$ from $\{0,1\}$ to $\boldsymbol{y}_0=2\boldsymbol{y}-1 \in \{-1,+1\}^{D\times H\times W}$, where $-1$ and $+1$ denote background and scar voxels, respectively.
Gaussian noise is then progressively added to $\boldsymbol{y}_0$ in the forward diffusion process for timestep $t \in \{1,\ldots,T\}$:
\begin{equation}
    q(\boldsymbol{y}_t \mid \boldsymbol{y}_0) = \mathcal{N}\!\left(\sqrt{\bar{\alpha}_t}\,\boldsymbol{y}_0,\,(1-\bar{\alpha}_t)\mathbf{I}\right)
\end{equation}
where $\bar{\alpha}_t = \prod_{s=1}^{t}(1-\beta_s)$ and $\{\beta_t\}_{t=1}^{T}$ is a linear noise schedule with $\beta_1=5\times10^{-4}$, $\beta_T=1.95\times10^{-2}$, $T=1000$.

In standard DDPM training, the denoising network is often parameterised to predict the injected Gaussian noise $\boldsymbol{\epsilon}$, which is effective for continuous image generation. Under this parameterisation, a noise-prediction network $\epsilon_\theta$ is trained to approximate $\boldsymbol{\epsilon}$ from the noisy sample, timestep, and condition:
\begin{equation}
    \epsilon_\theta(\boldsymbol{y}_t, t, \boldsymbol{x}) \approx \boldsymbol{\epsilon}
\end{equation}
and the clean sample is recovered indirectly as:
\begin{equation}
    \hat{\boldsymbol{y}}_0 =
    \frac{\boldsymbol{y}_t - \sqrt{1-\bar{\alpha}_t}\,\epsilon_\theta(\boldsymbol{y}_t,t,\boldsymbol{x})}
    {\sqrt{\bar{\alpha}_t}}
\end{equation}
For LA scar segmentation, however, the target is a sparse binary mask rather than a natural image, and the main objective is accurate voxel classification under severe class imbalance. In contrast to the conventional noise-prediction network $\epsilon_\theta$, we therefore parameterise CMDiff $g_\theta$ to directly predict the clean mask $\hat{\boldsymbol{y}}_0$ conditioned on the noisy mask $\boldsymbol{y}_t$ and the LGE image $\boldsymbol{x}$.
\begin{equation}
    \hat{\boldsymbol{y}}_{0,\theta}
    = g_\theta(\boldsymbol{y}_t,t,f_\phi(\boldsymbol{x}))
\end{equation}
where $f_\phi$ denotes the DCE image encoder introduced in Sec.~\ref{sec:dce}.
This parameterisation aligns the diffusion output with the segmentation target and permits direct supervision with the losses in Sec.~\ref{sec:loss}. We test it as a task-specific design hypothesis against noise prediction, for which the clean mask is recovered indirectly through the scheduler equation.

\subsection{Network architecture and image conditioning}

The denoising network combines the current noisy mask with anatomical and intensity information from LGE-MRI. A residual 3D U-Net-style backbone with sinusoidal timestep embeddings, denoted SinUNet3D~\cite{cciccek20163d,he2016deep}, receives $[\boldsymbol{y}_t,\boldsymbol{x}] \in \mathbb{R}^{2 \times D \times H \times W}$ and predicts $\hat{\boldsymbol{y}}_0$. The timestep embedding identifies the current noise level.

Because input concatenation alone does not provide LGE features at every resolution, we add a lightweight Dynamic Conditional Encoder (DCE), following prior diffusion segmentation work~\cite{wu2024medsegdiff}.
\label{sec:dce}
The DCE processes the LGE image $\boldsymbol{x}$ and produces feature maps at multiple spatial resolutions.
These features are combined with the timestep embedding to generate channel-wise scaling and shifting parameters, which modulate the normalised UNet features in a Feature-wise Linear Modulation (FiLM)-like manner~\cite{perez2018film}.
Modulation is applied at multiple UNet stages, and the FiLM output layers are zero-initialised so that the DCE initially behaves close to an identity transformation. The resulting model $g_\theta = (\text{SinUNet3D}, f_\phi)$ jointly adapts to LGE appearance and diffusion timestep.

\subsection{Training and inference}
\label{sec:loss}

Given the clean-mask target, training combines diffusion reconstruction with segmentation supervision. The diffusion component penalises the discrepancy between $\hat{\boldsymbol{y}}_0$ and $\boldsymbol{y}_0$.
We use a signal-to-noise-ratio (SNR)-weighted MSE loss~\cite{hang2023effdiff}:
\begin{equation}
    \mathcal{L}_\text{diff}(t) =
    \frac{\mathrm{SNR}(t)}{1+\mathrm{SNR}(t)}
    \left\|\hat{\boldsymbol{y}}_0 - \boldsymbol{y}_0\right\|_2^2
\end{equation}
where $\mathrm{SNR}(t) = \bar{\alpha}_t / (1-\bar{\alpha}_t)$.
This weighting reduces the influence of very noisy timesteps, where the clean mask is weakly recoverable, and gives greater emphasis to denoising states that provide more informative supervision.

The segmentation term applies soft Dice and weighted binary cross-entropy (BCE) to $\hat{p}=\sigma(\hat{\boldsymbol{y}}_0)$ to address foreground--background imbalance.
Because voxel-wise segmentation losses are most informative when the noisy mask remains sufficiently close to the clean target, we modulate this term by a timestep-dependent factor:
\begin{equation}
    \mathcal{L}_\text{seg}(t) = \left(1 - \frac{t}{T}\right)\left(\lambda_\text{Dice}\mathcal{L}_\text{Dice} + \lambda_\text{BCE}\mathcal{L}_\text{BCE}\right)
\end{equation}
The final objective combines the diffusion reconstruction term and the timestep-weighted segmentation term:
\begin{equation}
    \mathcal{L}(t) = \mathcal{L}_\text{diff}(t) + \lambda_\text{seg}\mathcal{L}_\text{seg}(t)
\end{equation}
Here, $(1-t/T)$ controls how strongly the segmentation loss is applied across diffusion timesteps, whereas $\lambda_\text{seg}$ controls the overall contribution of segmentation supervision relative to the diffusion loss.
In all experiments, we set $\lambda_\text{Dice}=0.8$, $\lambda_\text{BCE}=0.2$, and $\lambda_\text{seg}=0.5$.

At inference, CMDiff applies the reverse process to a random mask conditioned on the LGE image. We average $N=3$ independent soft predictions and binarise the result at zero. To assess the accuracy--cost trade-off, we compare DDIM~\cite{song2021ddim}, DPM-Solver++~\cite{lu2022dpmsolver}, Euler Discrete, and UniPC~\cite{zhao2023unipc} across different step counts.

\section{Experiments}
\label{sec:experiments}

\subsection{Dataset and preprocessing}
All experiments use the public LASCARQS 2022 dataset of 3D LGE-MRI from patients with AF~\cite{zhuang2023left}. Of 130 subjects with LA anatomy labels, 60 also have scar annotations. Accordingly, separate fixed five-fold partitions are used for the two tasks. Each scar fold contains 48 training and 12 held-out subjects, stratified by scar burden; each LA fold contains 104 and 26, respectively. Within each task, CMDiff and the deterministic baseline use identical held-out cases. Images are resampled to $0.625\times0.625\times2.5\,\mathrm{mm}^3$ and cropped around the LA to $60\times384\times256\times40$ voxels. Data augmentation by MONAI library~\cite{cardoso2022monai} comprises random bias field, elastic deformation, rotation, zoom, smoothing, contrast adjustment, intensity shift, and Gaussian noise. Patch size uses $288\times192\times32$.

\subsection{Implementation details}
Training is performed on an NVIDIA RTX A6000 GPU for 3000 epochs. AdamW uses weight decay $10^{-5}$, a 500-step linear warmup, and cosine annealing to $10^{-6}$. A 20-step DDIM sampler is used for validation and model selection~\cite{song2021ddim}; the broader scheduler benchmark is performed after training. An exponential moving average of the weights with decay 0.9999 is used for validation and inference.

\subsection{Evaluation metrics}
We report Dice similarity coefficient (DSC) for overlap, 95th-percentile Hausdorff distance (HD95) for worst-case boundary error, and average surface distance (ASD) for mean boundary error. Surface distances are calculated in millimetres using the resampled spacing; inference time is reported per subject. Unless stated otherwise, segmentation results are summarised as the mean $\pm$ standard deviation across held-out subjects.

\subsection{Stochastic prediction variability analysis}

To analyse within-model variability beyond the $N=3$ samples used for segmentation, we generate $K=20$ predictions per held-out case from independent Gaussian masks. Each clean-mask prediction $\hat{\boldsymbol{y}}^{(k)}_0 \in [-1,1]$ is mapped to $\boldsymbol{p}^{(k)}=(\hat{\boldsymbol{y}}^{(k)}_0+1)/2$, with mean $\bar{\boldsymbol{p}}=K^{-1}\sum_{k=1}^{K}\boldsymbol{p}^{(k)}$. The mean is thresholded at 0.5. Variance measures disagreement across predictions:
\begin{equation}
    \mathrm{Var}(\boldsymbol{p}) =
    \frac{1}{K}\sum_{k=1}^{K}\left(\boldsymbol{p}^{(k)}-\bar{\boldsymbol{p}}\right)^2
    \label{eq:uncertainty_variance}
\end{equation}
The corresponding standard deviation is $\mathrm{Std}(\boldsymbol{p})=\sqrt{\mathrm{Var}(\boldsymbol{p})}$. Binary entropy measures how close the mean prediction is to 0.5:
\begin{equation}
    \mathcal{H}(\bar{\boldsymbol{p}}) =
    -\bar{\boldsymbol{p}}\log_2(\bar{\boldsymbol{p}})
    -(1-\bar{\boldsymbol{p}})\log_2(1-\bar{\boldsymbol{p}})
    \label{eq:uncertainty_entropy}
\end{equation}
with probabilities clipped for numerical stability. For the deterministic baseline, $K=20$ predictions are obtained with Monte Carlo dropout. Both models use the same foreground-restricted entropy and variance summaries because whole-volume values are dominated by background. These quantities describe sampling dispersion; they are not calibrated error probabilities.

\section{Results}
\label{sec:results}

We first test the prediction target, then evaluate image conditioning and sampler choice, and finally compare CMDiff with the deterministic baseline at segmentation and scar-burden levels.

\subsection{Segmentation performance of CMDiff}

Table~\ref{tab:target_ablation} compares clean-mask and noise prediction under the same loss design, training schedule, DCE conditioning, and 20-step DDIM inference. The noise-prediction variant yields near-zero DSC and large surface distances, whereas clean-mask prediction reaches a DSC of 0.4675 and substantially lower boundary errors. We therefore use clean-mask prediction in the remaining experiments.

\vspace{-0.4cm}
\begin{table}[!htbp]
\centering
\caption{Clean-mask prediction vs. noise prediction under the matched implementation used in this study. Both models use DCE and DDIM with 20 inference steps.}
\label{tab:target_ablation}
\small
\begin{tabular*}{\textwidth}{@{\extracolsep{\fill}}lccc@{}}
\toprule
\textbf{Prediction target} & \textbf{DSC} $\uparrow$ & \textbf{HD95} $\downarrow$ & \textbf{ASD} $\downarrow$ \\
\midrule
Noise $\boldsymbol{\epsilon}$ & 0.0006 $\pm$ 0.0003 & 92.6167 $\pm$ 4.2491 & 52.6770 $\pm$ 3.7071 \\
Clean mask $\boldsymbol{y}_0$ & 0.4675 $\pm$ 0.0807 & 12.5934 $\pm$ 6.3477 & 1.9590 $\pm$ 0.9818 \\
\bottomrule
\end{tabular*}
\end{table}
\vspace{-0.4cm}

Having selected the prediction target, we next test whether multi-scale conditioning improves the denoiser. CMDiff contains 15.17 million trainable parameters: 13.80\,M in SinUNet3D and 1.37\,M in the DCE, a 9.0\% overhead. Compared with the same backbone using image concatenation alone, DCE improves DSC in most scheduler settings in Table~\ref{tab:scheduler_benchmark}. We then examine whether additional inference steps improve accuracy. For scar, UniPC at 50 steps gives the highest DSC (0.4687), whereas 20 steps remains close with lower runtime (Table~\ref{tab:scheduler_benchmark}). More steps do not consistently improve DSC. For LA, DPM-Solver++ at 20 steps gives the highest DSC (0.8709), and DPM-Solver++ and UniPC remain stable across step counts; DDIM degrades as the step count increases (Table~\ref{tab:la_scheduler_benchmark}).

\vspace{-0.6cm}

\begin{table}[H]
\centering
\captionsetup{font=scriptsize,skip=0pt}
\scriptsize
\setlength{\tabcolsep}{2pt}
\renewcommand{\arraystretch}{0.60}
\caption{Scar CMDiff scheduler benchmark with and without DCE across 5-fold cross-validation. $\Delta$DSC is computed as w/ DCE minus w/o DCE.}
\label{tab:scheduler_benchmark}
\begin{tabular*}{\textwidth}{@{\extracolsep{\fill}}llccccc@{}}
\toprule
\multirow{2}{*}{\textbf{Scheduler}} & \multirow{2}{*}{\textbf{Steps}} & \multicolumn{3}{c}{\textbf{DSC} $\uparrow$} & \multicolumn{2}{c}{\textbf{Time/case (s)} $\downarrow$} \\
\cmidrule(lr){3-5}\cmidrule(lr){6-7}
 & & \textbf{w/o DCE} & \textbf{w/ DCE} & \textbf{$\Delta$} & \textbf{w/o DCE} & \textbf{w/ DCE} \\
\midrule
\multirow{4}{*}{DDIM} 
 & 20  & 0.4595 $\pm$ 0.09 & 0.4675 $\pm$ 0.08 & +0.008 & 225 & 221 \\
 & 50  & 0.4596 $\pm$ 0.09 & 0.4686 $\pm$ 0.08 & +0.009 & 543 & 555 \\
 & 100 & 0.4549 $\pm$ 0.09 & 0.4649 $\pm$ 0.08 & +0.010 & 1182 & 1110 \\
 & 200 & 0.4381 $\pm$ 0.09 & 0.4237 $\pm$ 0.09 & -0.014 & 2009 & 2218 \\
\midrule
\multirow{4}{*}{DPM-Solver++} 
 & 20  & 0.4596 $\pm$ 0.09 & 0.4667 $\pm$ 0.08 & +0.007 & 226 & 222 \\
 & 50  & 0.4566 $\pm$ 0.09 & 0.4682 $\pm$ 0.08 & +0.012 & 562 & 555 \\
 & 100 & 0.4592 $\pm$ 0.09 & 0.4661 $\pm$ 0.08 & +0.007 & 1131 & 1110 \\
 & 200 & 0.4587 $\pm$ 0.09 & 0.4677 $\pm$ 0.08 & +0.009 & 2255 & 2220 \\
\midrule
\multirow{4}{*}{Euler Discrete}
 & 20  & 0.4349 $\pm$ 0.09 & 0.4588 $\pm$ 0.09 & +0.024 & 226 & 222 \\
 & 50  & 0.4339 $\pm$ 0.09 & 0.4589 $\pm$ 0.08 & +0.025 & 560 & 555 \\
 & 100 & 0.4339 $\pm$ 0.09 & 0.4586 $\pm$ 0.09 & +0.025 & 1128 & 1111 \\
 & 200 & 0.4325 $\pm$ 0.09 & 0.4581 $\pm$ 0.09 & +0.026 & 1482 & 2218 \\
\midrule
\multirow{4}{*}{UniPC}
 & 20  & \textbf{0.4606 $\pm$ 0.09} & 0.4662 $\pm$ 0.08 & +0.006 & 103 & 229 \\
 & 50  & 0.4596 $\pm$ 0.09 & \textbf{0.4687 $\pm$ 0.08} & +0.009 & 257 & 573 \\
 & 100 & 0.4569 $\pm$ 0.09 & 0.4668 $\pm$ 0.08 & +0.010 & 514 & 1145 \\
 & 200 & 0.4587 $\pm$ 0.09 & 0.4668 $\pm$ 0.08 & +0.008 & 1029 & 2288 \\
\bottomrule
\end{tabular*}

\caption{LA CMDiff with DCE scheduler benchmark across 5-fold cross-validation.}
\label{tab:la_scheduler_benchmark}
\begin{tabular*}{\textwidth}{@{\extracolsep{\fill}}llcc@{}}
\toprule
\textbf{Scheduler} & \textbf{Steps} & \textbf{DSC} $\uparrow$ & \textbf{Time/case (s)} $\downarrow$ \\
\midrule
\multirow{4}{*}{DDIM} 
 & 20  & 0.8697 $\pm$ 0.08 & 177 \\
 & 50  & 0.8628 $\pm$ 0.08 & 501 \\
 & 100 & 0.8329 $\pm$ 0.08 & 574 \\
 & 200 & 0.7425 $\pm$ 0.10 & 980 \\
\midrule
\multirow{4}{*}{DPM-Solver++} 
 & 20  & \textbf{0.8709 $\pm$ 0.08} & 206 \\
 & 50  & 0.8706 $\pm$ 0.08 & 513 \\
 & 100 & 0.8702 $\pm$ 0.08 & 1023 \\
 & 200 & 0.8706 $\pm$ 0.08 & 2045 \\
\midrule
\multirow{4}{*}{Euler Discrete}
 & 20  & 0.8500 $\pm$ 0.09 & 203 \\
 & 50  & 0.8443 $\pm$ 0.09 & 456 \\
 & 100 & 0.8420 $\pm$ 0.09 & 499 \\
 & 200 & 0.8403 $\pm$ 0.10 & 996 \\
\midrule
\multirow{4}{*}{UniPC}
 & 20  & 0.8700 $\pm$ 0.08 & 101 \\
 & 50  & 0.8706 $\pm$ 0.08 & 251 \\
 & 100 & 0.8704 $\pm$ 0.08 & 502 \\
 & 200 & 0.8706 $\pm$ 0.08 & 1002 \\
\bottomrule
\end{tabular*}
\end{table}

\vspace{-0.5cm}
Table~\ref{tab:uncertainty_summary} then compares CMDiff with the deterministic models on identical held-out cases. For scar, CMDiff has lower DSC (0.47 vs. 0.51), higher HD95 (12.89 vs. 10.86 mm), and higher ASD (1.91 vs. 1.58 mm). For LA, CMDiff has lower DSC (0.87 vs. 0.88) and higher ASD (2.07 vs. 1.74 mm), but lower HD95 (7.31 vs. 7.82 mm). Entropy and variance describe repeated-sample variability and are not accuracy metrics.
\begin{table}[htb]
\centering
\captionsetup{font=scriptsize,skip=0pt}
\scriptsize
\setlength{\tabcolsep}{2pt}
\renewcommand{\arraystretch}{1.00}
\caption{Segmentation performance and repeated-sample variability for the deterministic baselines and CMDiff on identical held-out cases.}
\label{tab:uncertainty_summary}
\resizebox{\textwidth}{!}{%
\begin{tabular}{llcccccc}
\toprule
\textbf{Task} & \textbf{Model} & \textbf{DSC} $\uparrow$ & \textbf{HD95} $\downarrow$ & \textbf{ASD} $\downarrow$ & \textbf{Sampling source} & \textbf{Entropy$_\mathrm{fg}$} & \textbf{Var$_\mathrm{fg}$} \\
\midrule
Scar & Deterministic segmentation~\cite{zhang2026progressive} & 0.51 $\pm$ 0.08 & 10.86 $\pm$ 5.39 & 1.58 $\pm$ 1.03 & MC dropout & 0.1157 & 9.32e-03 \\
Scar & CMDiff (ours, UniPC 50 steps) & 0.47 $\pm$ 0.08 & 12.89 $\pm$ 6.39 & 1.91 $\pm$ 0.96 & Diffusion sampling & 0.3046 & 2.82e-02 \\
\midrule
LA & Deterministic segmentation~\cite{zhang2026progressive} & 0.88 $\pm$ 0.07 & 7.82 $\pm$ 6.10 & 1.74 $\pm$ 1.30 & MC dropout & 0.0796 & 2.53e-04\\
LA & CMDiff (ours, DPM 20 steps) & 0.87 $\pm$ 0.08 & 7.31 $\pm$ 5.60 & 2.07 $\pm$ 1.84 & Diffusion sampling & 0.0575 & 6.20e-03 \\
\bottomrule
\end{tabular}
}
\end{table}

\subsection{Qualitative visualisation results}

The examples provide a visual counterpart to the quantitative results across favourable, intermediate, and challenging cases. In favourable views, both methods recover the dominant scar distribution and LA cavity. The differences become clearer for sparse scars and difficult boundaries, where small components may be missed and false-positive regions may appear away from the atrial wall. Repeated-sample entropy highlights locations where diffusion predictions vary.

\begin{figure}[H]
\centering
\captionsetup{font=scriptsize,skip=0pt}
\setlength{\tabcolsep}{0pt}
\renewcommand{\arraystretch}{0.74}
\newcommand{\qualimg}[1]{\includegraphics[width=0.170\textwidth]{#1}}
\newcommand{\uncimg}[2]{%
\IfFileExists{#1}{\includegraphics[width=0.170\textwidth]{#1}}{%
\fbox{\parbox[c][0.09\textheight][c]{0.16\textwidth}{\centering\scriptsize #2}}}}
\newcommand{\quallabel}[1]{\scriptsize #1}
\begin{tabular*}{\textwidth}{@{\extracolsep{\fill}}ccccc@{}}
\qualimg{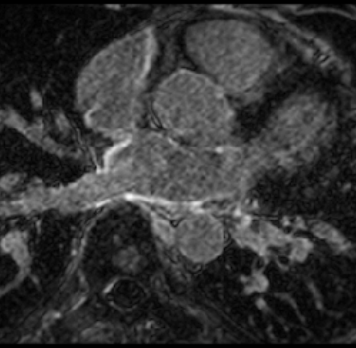} &
\qualimg{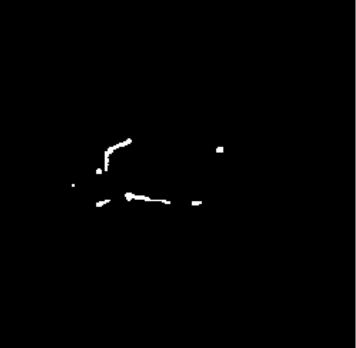} &
\qualimg{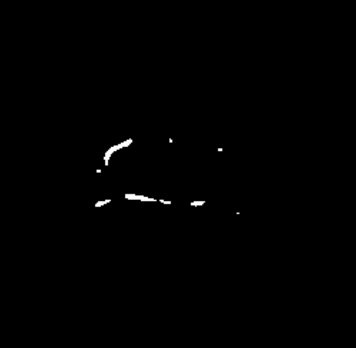} &
\qualimg{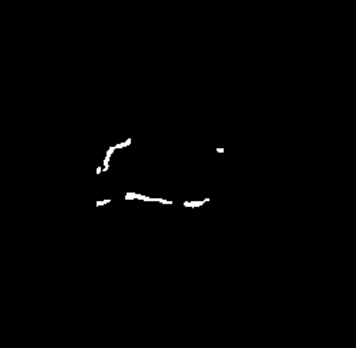} &
\uncimg{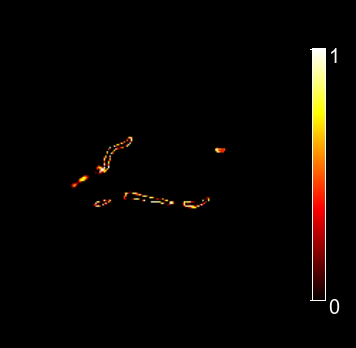}{scar\_06 entropy} \\
\quallabel{LGE-MRI} & \quallabel{Ground truth} & \quallabel{Seg (DSC=0.65)} & \quallabel{CMDiff (DSC=0.61)} & \quallabel{Entropy (0.2629)} \\[0.1em]
\qualimg{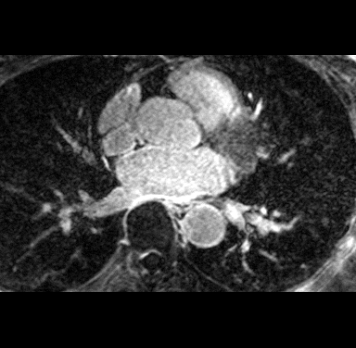} &
\qualimg{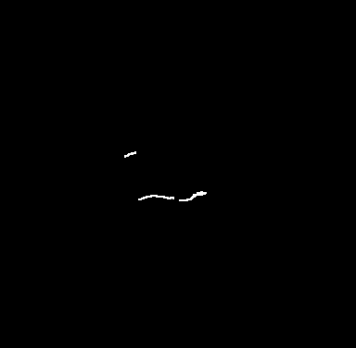} &
\qualimg{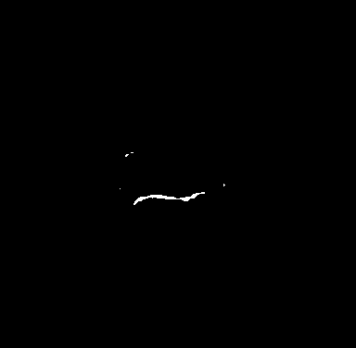} &
\qualimg{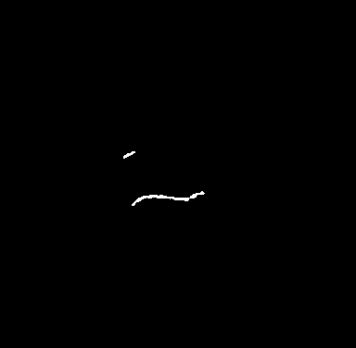} &
\uncimg{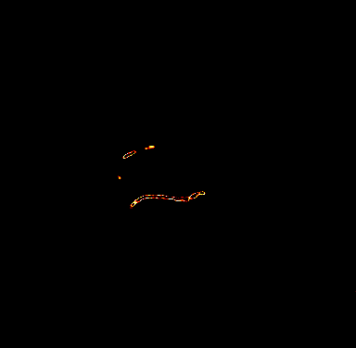}{scar\_51 entropy} \\
\quallabel{LGE-MRI} & \quallabel{Ground truth} & \quallabel{Seg (DSC=0.51)} & \quallabel{CMDiff (DSC=0.49)} & \quallabel{Entropy (0.2838)} \\[0.1em]
\qualimg{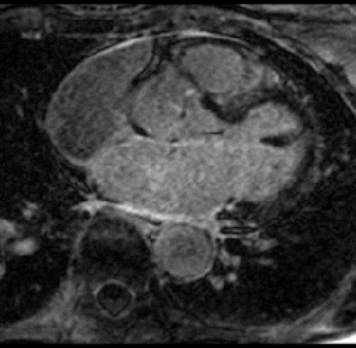} &
\qualimg{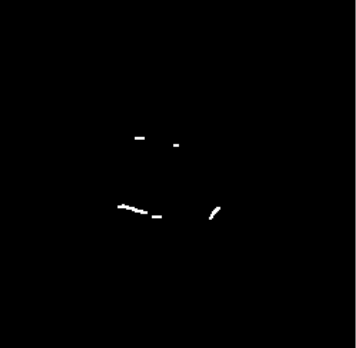} &
\qualimg{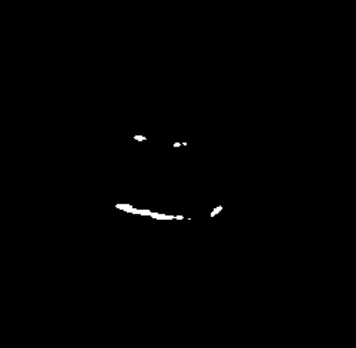} &
\qualimg{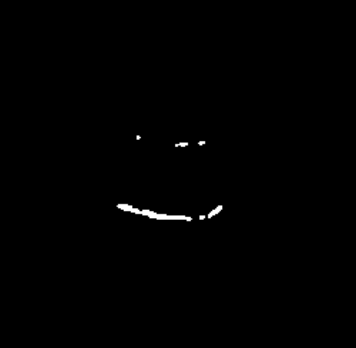} &
\uncimg{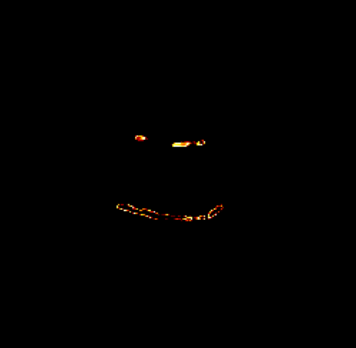}{scar\_14 entropy} \\
\quallabel{LGE-MRI} & \quallabel{Ground truth} & \quallabel{Seg (DSC=0.40)} & \quallabel{CMDiff (DSC=0.38)} & \quallabel{Entropy (0.2990)} \\[0.1em]
\qualimg{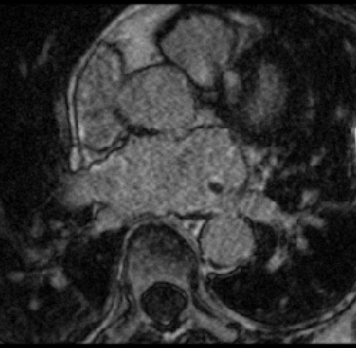} &
\qualimg{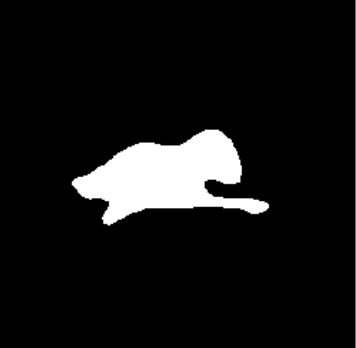} &
\qualimg{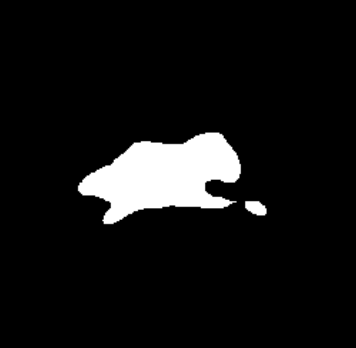} &
\qualimg{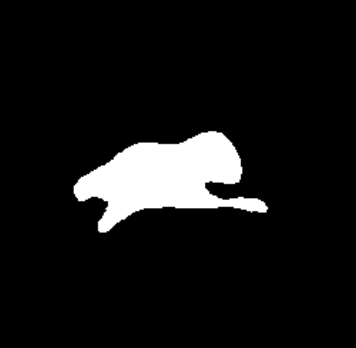} &
\uncimg{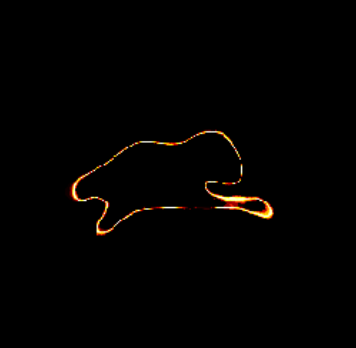}{LA\_85 entropy} \\
\quallabel{LGE-MRI} & \quallabel{Ground truth} & \quallabel{Seg (DSC=0.93)} & \quallabel{CMDiff (DSC=0.94)} & \quallabel{Entropy (0.0562)} \\[0.1em]
\qualimg{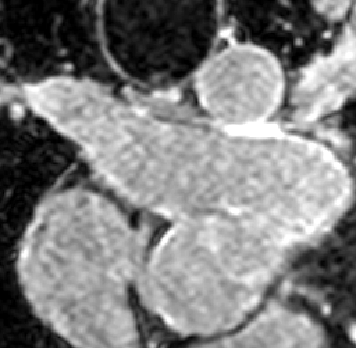} &
\qualimg{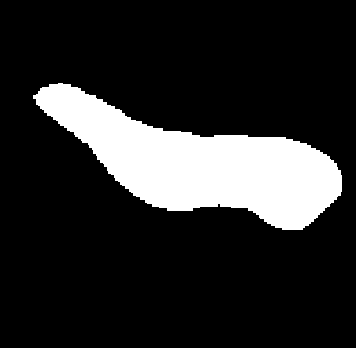} &
\qualimg{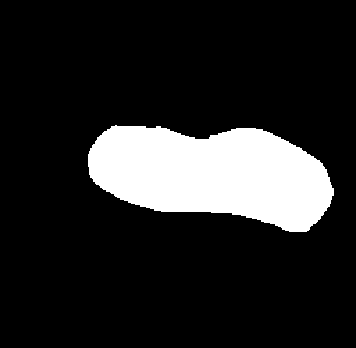} &
\qualimg{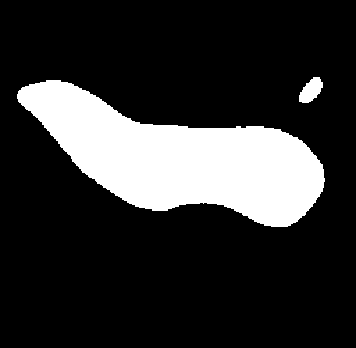} &
\uncimg{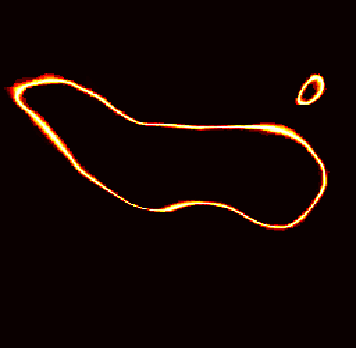}{LA\_12 entropy} \\
\quallabel{LGE-MRI} & \quallabel{Ground truth} & \quallabel{Seg (DSC=0.79)} & \quallabel{CMDiff (DSC=0.78)} & \quallabel{Entropy (0.0884)} \\[0.1em]
\qualimg{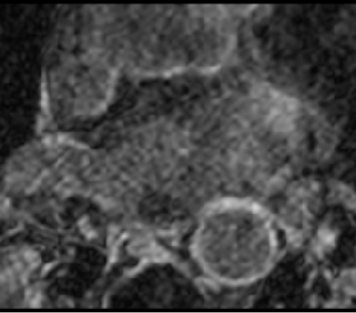} &
\qualimg{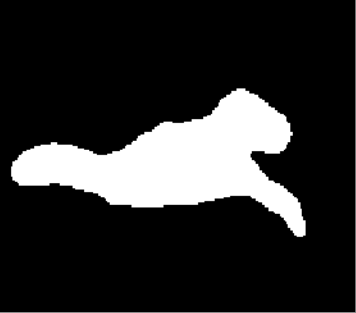} &
\qualimg{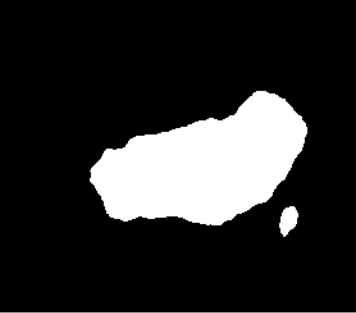} &
\qualimg{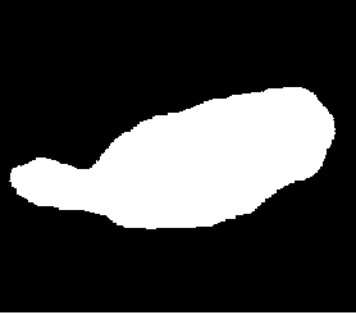} &
\uncimg{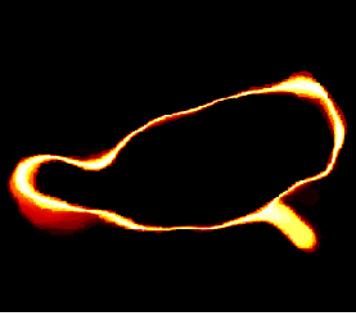}{LA\_34 entropy} \\
\quallabel{LGE-MRI} & \quallabel{Ground truth} & \quallabel{Seg (DSC=0.78)} & \quallabel{CMDiff (DSC=0.60)} & \quallabel{Entropy (0.1238)} \\[0.1em]
\end{tabular*}
\caption{Qualitative comparison between deterministic segmentation and CMDiff for representative scar and LA cases. From left to right: LGE-MRI image, ground truth, deterministic segmentation, CMDiff prediction, and CMDiff repeated-sample entropy map.}
\label{fig:qualitative_cases}
\end{figure}

\clearpage

\subsection{Scar burden biomarker evaluation}

To determine whether voxel-wise errors altered global quantification, scar burden was calculated as the scar-voxel count divided by a common 2.5\,mm LA-wall shell derived from the ground-truth cavity. Using the same denominator for all methods isolates differences in predicted scar extent.

\begin{figure}[H]
\centering
\captionsetup{font=footnotesize,skip=1pt}
\begin{subfigure}{0.90\textwidth}
\centering
\includegraphics[width=\textwidth]{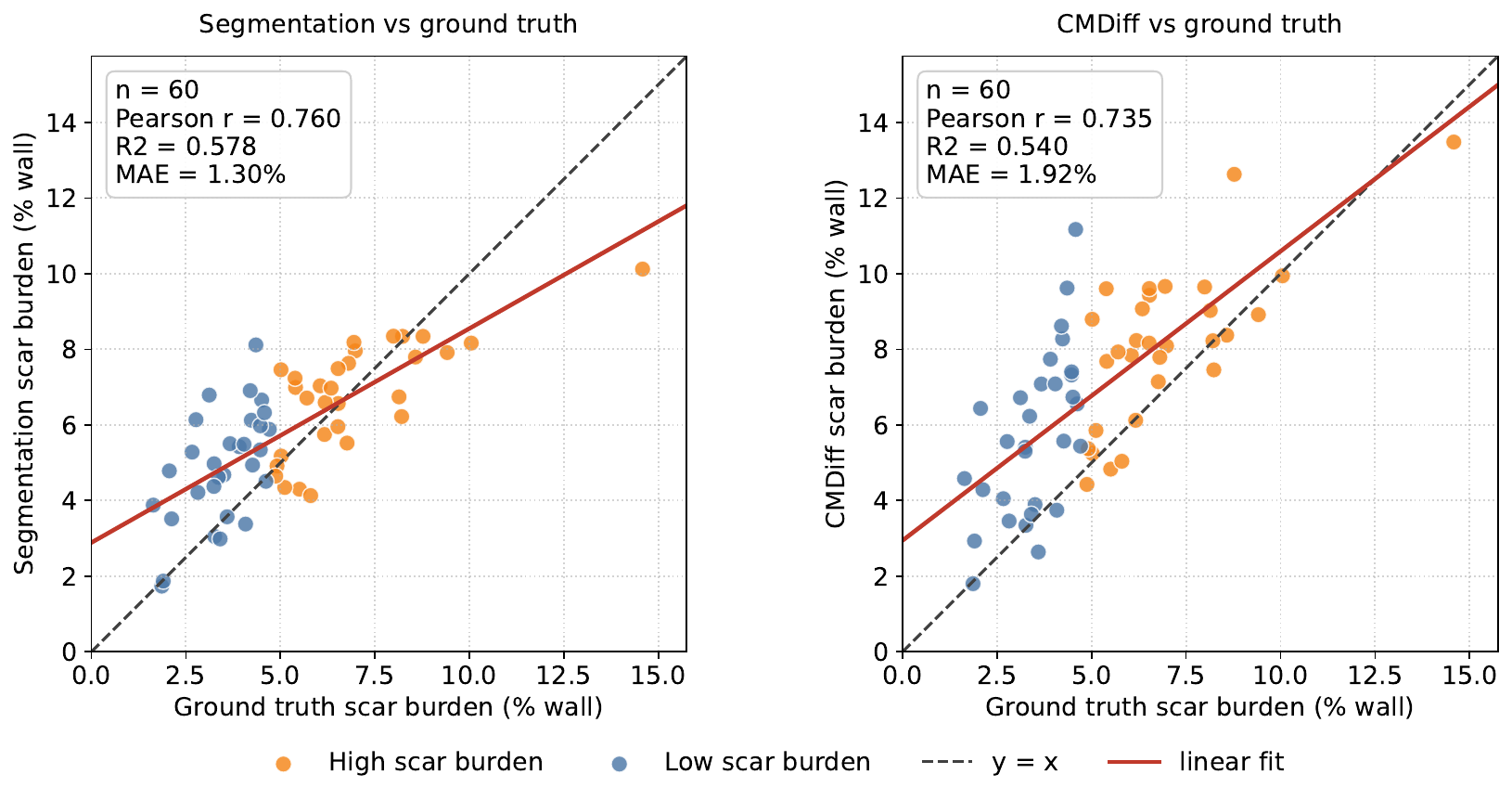}
\caption{Correlation between predicted and ground-truth scar burden.}
\label{fig:scar_burden_corr}
\end{subfigure}

\vspace{0pt}
\begin{subfigure}{0.90\textwidth}
\centering
\includegraphics[width=\textwidth]{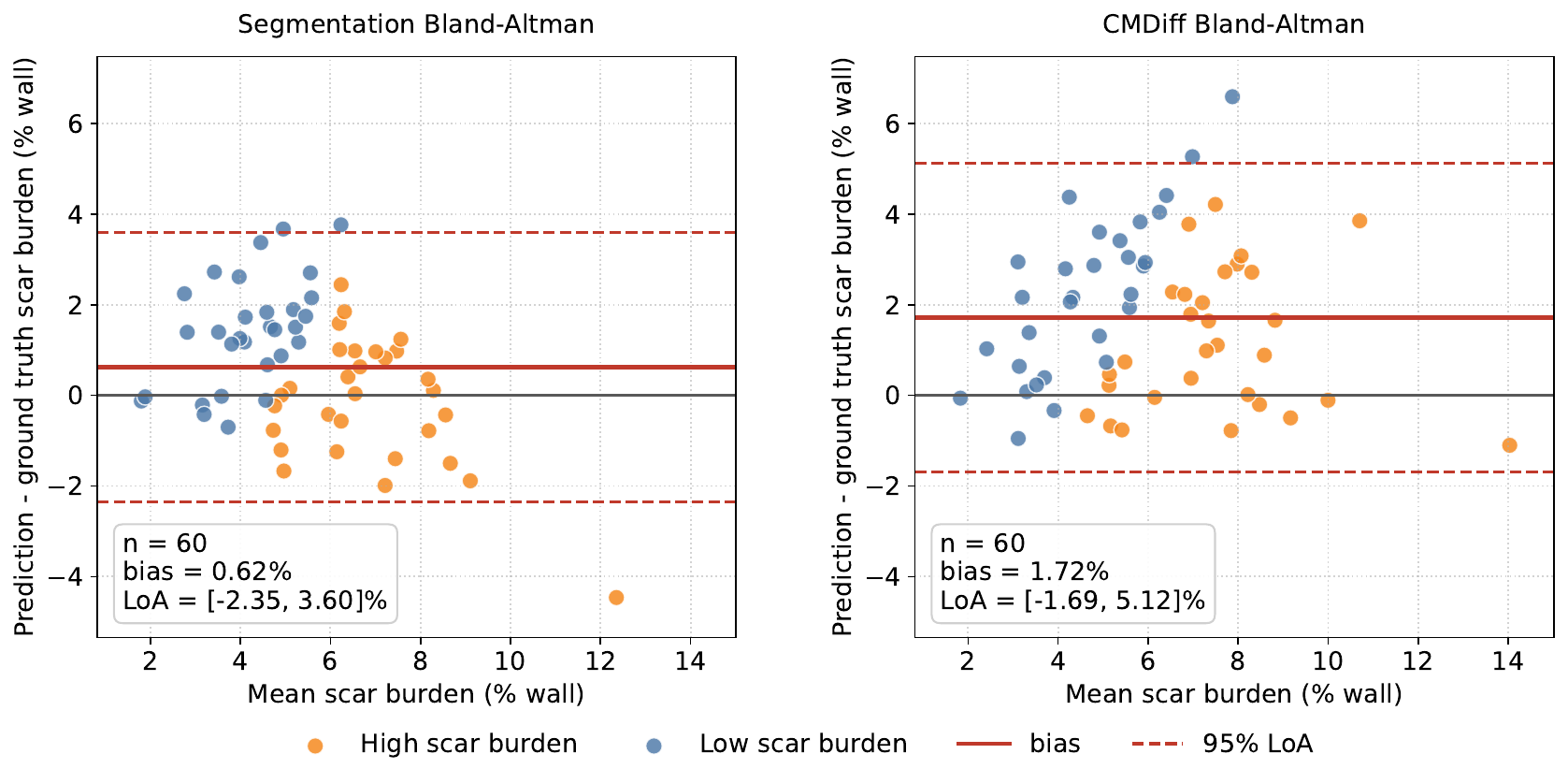}
\caption{Bland--Altman analysis of scar-burden agreement.}
\label{fig:scar_burden_ba}
\end{subfigure}
\caption{Scar-burden agreement between predicted masks and ground truth. In each subfigure, the deterministic segmentation result is shown on the left and CMDiff on the right.}
\label{fig:scar_burden}
\end{figure}

The deterministic baseline reaches Pearson $r=0.760$, $R^2=0.578$, and MAE $=1.30\%$ of wall volume (Fig.~\ref{fig:scar_burden}\subref{fig:scar_burden_corr}); its bias is $0.62\%$, with 95\% limits of agreement from $-2.35\%$ to $3.60\%$ (Fig.~\ref{fig:scar_burden}\subref{fig:scar_burden_ba}). CMDiff also follows the reference burden ($r=0.735$, $R^2=0.540$), but its larger MAE of $1.92\%$ indicates weaker agreement. Thus, both methods preserve the case-level trend, while the deterministic model is more accurate; this analysis does not establish clinical validity.

\clearpage

\section{Discussion}
\label{sec:discussion}

\paragraph{Main findings.}
The experiments answer the feasibility question in three steps. First, direct clean-mask prediction enables CMDiff to learn scar masks, whereas the tested noise-prediction configuration fails. Second, multi-scale LGE conditioning usually improves DSC, but additional sampling steps do not consistently improve accuracy. Third, CMDiff remains worse than the matched deterministic baseline in scar overlap, both surface-distance metrics, burden agreement, and inference cost. Conditional diffusion is therefore feasible, but not a performance advance in this setting.

\paragraph{Prediction target and conditioning.}
Clean-mask prediction connects each reverse step directly to Dice and BCE supervision, which may be advantageous for a sparse binary target. Its large gain over noise prediction nevertheless remains specific to the tested loss, timestep weighting, optimisation, and scheduler configuration; the noise-prediction baseline was not independently tuned. Once the target is fixed, the generally positive DCE effect indicates that multi-scale image features help the denoiser use LGE appearance beyond input concatenation. The remaining HD95 errors show that stronger conditioning alone does not enforce anatomically valid scar locations.

\paragraph{Sampling efficiency.}
The scheduler comparison shows that 20--50 reverse steps are sufficient for the best observed performance, while longer trajectories add cost without reliable accuracy gains. Faster solvers therefore reduce, but do not remove, the practical disadvantage of iterative inference relative to one-pass deterministic segmentation.

\paragraph{Relation to deterministic methods.}
The matched deterministic model uses LA anatomy as a geometric prior and performs better overall~\cite{zhang2026progressive}. This result is consistent with reports that distance maps, wall restrictions, and boundary-aware supervision benefit scar segmentation~\cite{zhang2023boundary,kundu2026twostage}. Although published scores are not directly comparable across protocols, they indicate that explicit anatomy is a more important current lever than stochastic sampling. The burden analysis leads to the same conclusion: CMDiff preserves the reference trend, but the deterministic model has tighter agreement.

\paragraph{Interpreting repeated samples.}
CMDiff naturally produces multiple predictions, allowing entropy and variance to quantify variation across repeated samples. In this study, these metrics are used to compare within-model sampling stability: lower values indicate more consistent outputs for the same image and inference settings. They do not measure segmentation accuracy and are therefore interpreted together with Dice and surface-distance metrics.

\paragraph{Limitations and future work.}
Three limitations define the scope of these findings. First, 60 scar-labelled subjects from one public dataset cannot establish generalisation across centres, scanners, or annotation styles. Second, iterative inference remains substantially slower than a deterministic forward pass. Third, CMDiff was trained only on the 60 scar-labelled cases and did not exploit the 70 additional LA-labelled cases through anatomy pretraining or shared representation learning; nor did it explicitly constrain scar predictions to the LA wall. The present results therefore do not determine whether anatomy-assisted diffusion would perform better. Future work should include multi-centre evaluation, independently tuned target parameterisations, faster sampling, and explicit anatomical conditioning. The 70 LA-only cases could first support pretraining of an LA encoder or wall representation. With a larger set of high-quality scar annotations, diffusion could then be evaluated as a refiner conditioned on the LGE image, LA anatomy, and a strong deterministic scar-probability map. In this formulation, diffusion would learn corrections to an initial segmentation rather than generate the complete scar mask from noise.

\section{Conclusion}
\label{sec:conclusion}

This feasibility study explored 3D conditional diffusion for left atrial scar segmentation from LGE-MRI. CMDiff learned to generate scar masks using clean-mask prediction, but it did not outperform the matched deterministic baseline in accuracy or efficiency. The moderate overlap and remaining surface errors of both approaches underscore the difficulty of delineating sparse, low-contrast scar along the thin LA wall. The study therefore establishes the technical feasibility of direct mask diffusion and identifies its current failure modes, motivating future diffusion refinement with larger, high-quality scar-labelled cohorts.

\section*{Acknowledgements}
This research is funded by Institut Universitaire d’Ingénierie en Santé (IUIS).


\bibliographystyle{unsrt}
\bibliography{reference}

\end{document}